\documentclass[review,authoryear]{elsarticle}

\usepackage[margin=2.5cm]{geometry}
\usepackage{lineno,hyperref}
\modulolinenumbers[5]
\usepackage[ruled,vlined]{algorithm2e}
\usepackage[utf8]{inputenc} 
\usepackage[T1]{fontenc}    
\usepackage{amsfonts}       
\usepackage{nicefrac}       
\usepackage{microtype}      
\usepackage{amsmath}
\usepackage[dvipsnames]{xcolor}
\usepackage{float}
\usepackage{graphicx}

\journal{Medical Image Analysis}
\begin{document}

\begin{frontmatter}

\title{Simple statistical methods for unsupervised brain anomaly detection on MRI are competitive to deep learning methods}

\author[nrad]{Victor Saase MD MSc\corref{cor1}}
\ead{victor.saase@uni-heidelberg.de}
\cortext[cor1]{Corresponding author}
\author[nrad]{Holger Wenz MD}
\author[biomed]{Thomas Ganslandt MD}
\author[nrad]{Christoph Groden MD}
\author[nrad,biomed]{Máté E. Maros MD MSc\corref{cor2}}
\cortext[cor2]{Clinical corresponding author}
\ead{maros@uni-heidelberg.de}

\address[nrad]{Departement of Neuroradiology, 
  Medical Faculty Mannheim of Heidelberg University, Theodor-Kutzer-Ufer 1-3, D-68167 Mannheim}
\address[biomed]{Departement of Biomedical Informatics at the Center for Preventive Medicine and Digital Health (CPD-BW), 
  Medical Faculty Mannheim of Heidelberg University, Theodor-Kutzer-Ufer 1-3, D-68167 Mannheim}

\begin{abstract}

Statistical analysis of magnetic resonance imaging (MRI) can help radiologists to detect pathologies that are otherwise likely to be missed.
Deep learning (DL) has shown promise in modeling complex spatial data for brain anomaly detection. However, DL models have major deficiencies: they need large amounts of high-quality training data, are difficult to design and train and are sensitive to subtle changes in scanning protocols and hardware. Here, we show that also simple statistical methods such as voxel-wise (baseline and covariance) models and a linear projection method using spatial patterns can achieve DL-equivalent (3D convolutional autoencoder) performance in unsupervised pathology detection. All methods were trained (N=395) and compared (N=44) on a novel, expert-curated multiparametric (8 sequences) head MRI dataset of healthy and pathological cases, respectively. We show that these simple methods can be more accurate in detecting small lesions and are considerably easier to train and comprehend. 
The methods were quantitatively compared using AUC and average precision and evaluated qualitatively on clinical use cases comprising brain atrophy, tumors (small metastases) and movement artefacts. Our results demonstrate that while DL methods may be useful, they should show a sufficiently large performance improvement over simpler methods to justify their usage. Thus, simple statistical methods should provide the baseline for benchmarks. Source code and trained models are available on GitHub (\url{https://github.com/vsaase/simpleBAD}). 
\end{abstract}


\end{frontmatter}

\begin{figure}[H]
    \centering
    \includegraphics[scale=0.4]{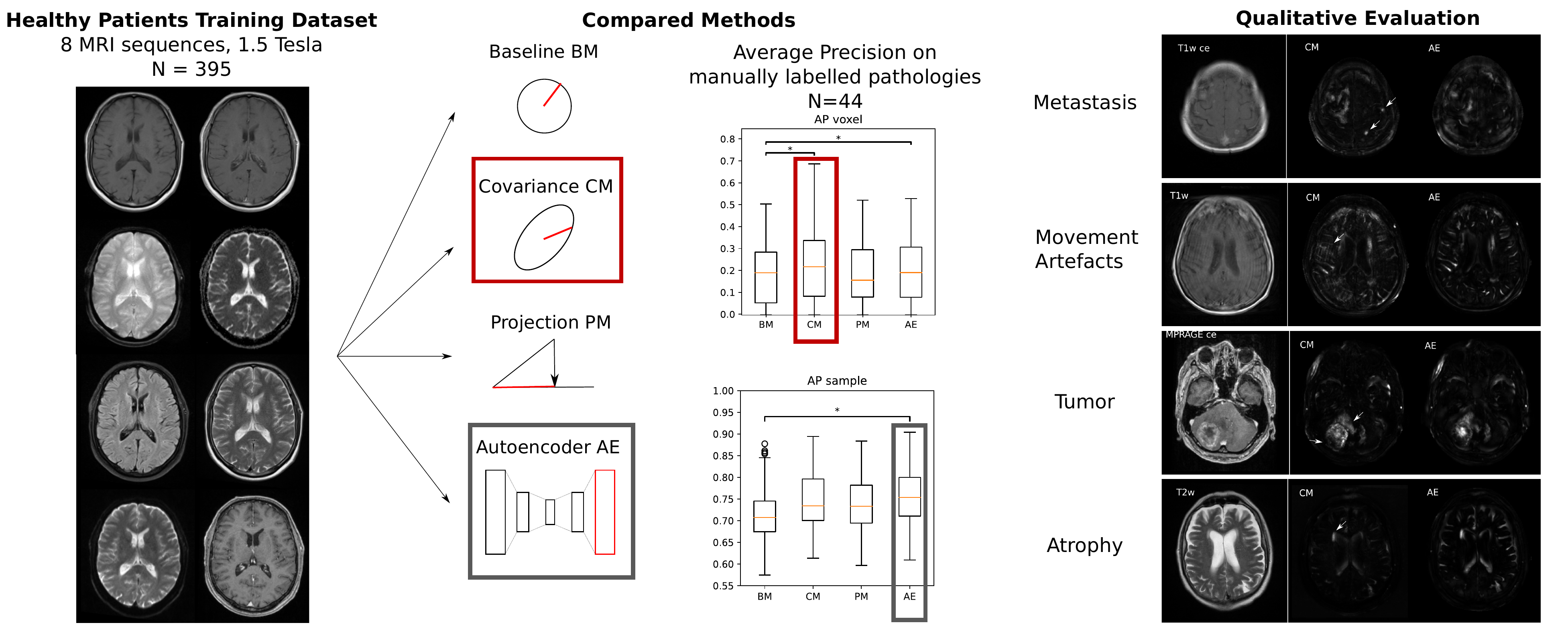} 
    \caption{Graphical Abstract}
    \label{fig:graphicalabstract}
\end{figure}


\section{Highlights} 
\begin{itemize}
    \item Unsupervised modeling of MR images can help radiologists detect pathologies.
    \item We provide a new expert-labelled multiparametric head MRI dataset (N=439).
    \item Voxel-wise covariance estimation is competitive to deep learning methods.
    \item Spatial models like autoencoders might not be sensitive to small lesions.
    \item Deep learning researchers should compare performance of novel approaches to simple statistical methods.
    
\end{itemize}

\section{Introduction}

The need for large scale annotated, manually segmented and site specific datasets in classical supervised statistical modeling and deep learning (DL) has limited the use of many applications in clinical practice. 
Thus, a line of work has been investigating the use of unsupervised DL methods in neuroradiology, specifically on brain MRI \citep{Schlegl2017,Baur2019,CHEN2020101713}. These unsupervised methods can be trained on healthy appearing MRI scans, which are more readily available than manually segmented scans \citep{Zimmerer2018ceVAE, Zimmerer2019}. Another advantage of unsupervised methods is that they are not biased towards the frequency of appearance of certain kinds of pathologies in the training dataset \citep{Maros2020machine}. 

A comprehensive comparison of unsupervised deep learning autoencoder models was recently published by \citet{Baur2020}. This study concluded that dense variational autoencoders should be preferred while no clear benefit was observed for generative adversarial networks (GAN)-based models and combinations or modifications of the standard variational autoencoders (VAE). However, in a more recent paper, the same authors showed competitive results between variants of basic non-variational autoencoders and VAEs \citep{Baur2020b}. In this light and because of their more stable training behavior, we considered a basic dense 3D convolutional autoencoder (AE) as the representative model of DL approaches. 

However, the performance of these complex DL-based models are seldom compared to simple statistical methods such as voxel-wise Gaussian distributions and projection methods equivalent to principal component analysis (PCA). Although voxel-wise distribution models are often used effectively together with clustering techniques for brain segmentation or lesion detection \citep{arnaud2018fully, ji2012fuzzy}. Using the PCA method, we achieved competitive results to DL models on the brain MRI-dataset of the Medical Out-of-Distribution Analysis Challenge (MOOD) at the 23rd International Conference on Medical Image Computing and Computer Assisted Intervention (MICCAI 2020). The MOOD challenge was held with the goal to improve upon the state-of-the-art and to tackle vulnerabilities of unsupervised anomaly detection on medical images \citep{mood_2020_3784230}.

Therefore in this study, we compared a basic dense autoencoder with voxel-wise Gaussian distributions and PCA on a novel expert-curated brain MRI dataset including 8 sequences of 395 healthy and 44 pathological cases. We show that these simple statistical methods can outperform the complex and computationally more demanding DL approach. Also, we provide detailed quantitative (including sample- and voxel-level evaluations) and qualitative neuroradiological comparisons of the investigated algorithms when detecting various clinical anomalies such as brain atrophy, large solitary or multiple small tumors or movement artefacts. Our results emphasize the importance of comparing complex DL-based methods and their performances against basic linear techniques when evaluating algorithms for anomaly detection on brain MRI.

\section{Methods}

\subsection{Dataset} 
We built a database of all head MRI (1.5 Tesla) studies performed between 10/2011-12/2019 at our institution and selected studies that contained T1-weighted (T1w), T2w, fat saturated FLAIR, DWI (ADC + b0 TRACE + b1000 TRACE), T2* and contrast enhanced T1w (T1w ce) and MPRAGE sequences. The study was approved by the local ethics committee (Medical Ethics Commission II, Medical Faculty Mannheim, Heidelberg University, approval nr.: 2017-825R-MA, 2017-828R-MA). Because of the retrospective nature of the study written informed consents were waived by the ethics committee. 
The corresponding radiological reports for the obtained MRI studies were extracted and manually classified as healthy vs. pathological. This resulted in a set of 5076 studies. In a second step, all studies with reports indicating healthy results were visually inspected for any missed or ambiguous pathologies, resulting in a final curated dataset of 395 healthy studies. These were further split randomly into a training set of 351 (88.9\%) healthy studies and a testing set of 44 (11.1\%) healthy studies. Additionally, we randomly sampled 44 pathological studies and manually segmented all pathological voxels with ITK-SNAP (\url{http://www.itksnap.org/pmwiki/pmwiki.php}) \citep{py06nimg}.

For every study, the sequences were rigidly co-registered to the T2w sequence, which in turn was affinely registered to a standard head template in MNI space, using the ANTs toolkit (\url{http://stnava.github.io/ANTs/}) \citep{Avants2014}. After linear interpolation and caudal and cranial cropping this resulted in data arrays of size 9x192x224x192 (sequences x width x depth x height) with isotropic voxel size of 1x1x1mm. 

The last preprocessing step was to normalize every sequence by subtracting its mean value and dividing it by the standard deviation, both computed over the voxels in the MNI head mask. Bias field correction was not used in this study as it is computationally expensive and can, especially with large regions of edema, lead to wrong results when applied to pathological scans. Empirically, the MR images of our institution had low bias fields compared to other openly available datasets.

\subsection{Investigated methods}
In the following subsections, we describe the investigated methods including a voxel-wise baseline model (BM; \ref{Baseline model}), a  voxel-wise covariance model (CM; \ref{Covariance model}), a linear projection method using spatial patterns (PM; \ref{Projection method}); and a DL-based approach  using a 3D convolutional autoencoder (AE; \ref{Deep learning method}). All methods were implemented in Python 3.8 and PyTorch 1.7.

\subsubsection{Baseline model} \label{Baseline model}
For the baseline model (BM), voxel- and sequence-wise z-transformation was done at every voxel, which is equivalent to estimating an uncorrelated multivariate normal distribution at every voxel \citep{gelman:2006,hastie2009elements}. 
Thus, the vector norm of the z-maps, that corresponds to the distance to the mean of the normal distribution at every voxel, becomes the measure of pathology.

Let $x_{isv}$ be the (preprocessed) value of the $i$-th training sample at sequence $s$ (i.e. channel) and voxel $v$. Then

\begin{align*}
\mu_{sv} &= \sum_{i=0}^N x_{isv} / N \\
\sigma_{sv} &= \sqrt{\sum_{i=0}^N (x_{isv}-\mu_{sv})^2 / (N-1)} \\
z_{isv} &= (x_{isv}-\mu_{sv})/\sigma_{sv}
\end{align*}
    
In the following, we denote the normalized data samples by $\mathbf{x}$, the voxel mean corrected data by $\mathbf{y}$ and the voxel-wise z-transformed data by $\mathbf{z}$.

\subsubsection{Covariance model} \label{Covariance model}
The voxel-wise covariance between the input sequences was estimated at every voxel, resulting in a correlated multivariate normal distribution for every voxel \citep{gelman:2006,hastie2009elements}. 
For this voxel-wise covariance model (CM), the measure of pathology was the Mahalanobis distance, which is the generalization of the distance to the mean in a correlated multivariate normal distribution \citep{mahalanobis1936generalized}.

Let $\mathbf{Y}_v$ be the NxS matrix with elements $(y_{isv})_{is} = (x_{isv} - \mu_{sv})_{is}$ for every voxel $v$ and $\mathbf{y}_{iv}$ be the S-dimensional vector with elements $(x_{isv} - \mu_{sv})_s$. Then the Mahalanobis distance $r_{iv}$ is defined by

\begin{align*}
\mathbf{\Sigma}_v &= \mathbf{Y}_v^\top \cdot \mathbf{Y}_v / (N-1) \\
r_{iv} &= \mathbf{y}_{iv}^\top \cdot \mathbf{\Sigma}_v^{-1} \cdot \mathbf{y}_{iv}
\end{align*}

\subsubsection{Projection method} \label{Projection method}
For the linear projection method (PM), the $\mathbf{z}$-maps of the baseline model (\ref{Baseline model}) were used to build an orthogonal basis of the S*V-dimensional vector space (S number of sequences, V number of voxels) spanned by the healthy training dataset. This is equivalent to performing principal component analysis (PCA) over all voxels and also equivalent to Gaussian Process regression at every voxel with a linear kernel over all voxels \citep{10.5555/646257.685385}. Hence, we could incorporate global spatial patterns into the model. 
Here, the measure of pathology was the norm of the residual, which was obtained by subtracting the projection on the healthy subspace of the input vector from itself.

\begin{figure}[ht]
  \centering
  \begin{minipage}{.7\linewidth}
    \begin{algorithm}[H]
    \SetAlgoLined
    \KwResult{Residual voxel vector $\mathbf{r}$}
     input: $\mathbf{z}_i \in \mathbb{R}^{S*V}$ training vectors for $i \in [1,..,N]$, $\mathbf{z} \in \mathbb{R}^{S*V}$ test vector\;
     $\mathbf{r} \leftarrow \mathbf{z}$\;
     \For{i=1 \KwTo N}{
      $c \leftarrow \mathbf{r} \cdot \mathbf{z}_i / |\mathbf{z}_i|$ \;
      $\mathbf{r} \leftarrow \mathbf{r} - c * \mathbf{z}_i / |\mathbf{z}_i|$\;
     }
     output: $\mathbf{r}$
     \caption{Projection method}
    \end{algorithm}
  \end{minipage}
\end{figure}

\subsubsection{Deep learning method} \label{Deep learning method}
We trained a 3D convolutional autoencoder (AE) on the 
$\mathbf{z}$-maps of the baseline model. Training on $\mathbf{z}$, instead of $\mathbf{x}$, is conceptually similar to batch normalization 
before the first layer and reverting after the last layer \citep{ioffe2015batch}. The voxels outside the MNI head mask are set to zero. The model is expected to learn only the healthy spatial distribution of the inputs. In this case, the measure of pathology was the norm of the residual obtained by subtracting the prediction from the input.

The architecture of the autoencoder is described in Table \ref{autoencoderarch}. It was inspired by the winning architecture of the MICCAI Brain Tumor Segmentation (BraTS) Challenge 2018 \citep{Myronenko2018} and by a large comparison of autoencoder models for unsupervised segmentation \citep{Baur2020}. We used skip connections over convolutional layers for faster training. Downsampling and upsampling was performed by convolutional layers and additionally by linear interpolation skip connections for the first 9 layers, corresponding to the number of input sequences \citep{Myronenko2018,Baur2020}.

The autoencoder was trained with the Adam optimization algorithm \citep{kingma2014adam} with L2 loss, batch size 1, learning rate 1e-4 and weight decay of 1e-3. Training was stopped when we did not observe further decrease of the testing loss for 25 epochs or a maximum of 70 epochs. Then the model with the lowest testing loss was selected, for details see Supplementary materials (Figure \ref{fig:supplconvergence}). 

\begin{table}[H]
\begin{center}
\begin{tabular}{ l l l l }
 Layer & Operation & Input Dimensions & Output Dimensions  \\
 \hline  \\ 
 1 & 4x4x4 Convolution with Downsampling Skip & 9x192x224x192 & 32x96x112x96 \\ 
 2 & Double 3x3x3 Convolution with Skip &  &  \\
 3 & 4x4x4 Convolution with Downsampling Skip & 32x96x112x96 & 64x48x56x48 \\ 
 4 & Double 3x3x3 Convolution with Skip &  &  \\ 
 5 & 4x4x4 Convolution with Downsampling Skip & 64x48x56x48 & 16x24x28x24 \\ 
 6 & Double 3x3x3 Convolution with Skip &  &  \\ 
 7 & Bottleneck Fully Connected & 16x24x28x24 & 512 \\ 
 8 & Bottleneck Fully Connected & 512 & 16x24x28x24 \\ 
 9 & 3x3x3 Convolution with Skip &  &  \\ 
 10 & 4x4x4 Transposed Convolution with Upsampling Skip & 16x24x28x24 & 64x48x56x48\\ 
 11 & 3x3x3 Convolution with Skip &  &  \\ 
 12 & 4x4x4 Transposed Convolution with Upsampling Skip & 64x48x56x48 & 32x96x112x96\\ 
 13 & 3x3x3 Convolution with Skip &  &  \\ 
 14 & 4x4x4 Transposed Convolution with Upsampling Skip & 32x96x112x96 & 9x192x224x192\\ 
 \hline  \\ 
\end{tabular}
\end{center}
\caption{Architecture of the convolutional autoencoder. After every convolution and after the fully connected layers a leaky rectified linear unit (ReLU) transfer function with parameter 0.2 was applied. The additive skip connections bypass the convolution and the transfer function. The down- and upsampling skip connections used linear interpolation on the first 9 channels of their input and add to the first 9 channels of the output layers. The 4x4x4 convolution layers performed upscaling and downscaling with stride 2. All convolutional layers used padding of 1.}
\label{autoencoderarch}
\end{table}

\subsubsection{Performance evaluation}
The algorithms were quantitatively evaluated using sample- and voxel-based comparisons \citep{mood_2020_3784230}. For the sample-based analysis, we used the sum of the voxel-wise scores as the sample score. We evaluated the methods by computing the average precision (AP) and the area under the Receiver Operating Characteristic Curve (AUC). 
For the sample-based analyses, we used the bootstrap method to estimate the distributions of the metrics and to compare the methods \citep{Efron1979}. Briefly, 100000 random bootstrap samples were drawn from the dataset with replacement. The AP and AUC metrics were calculated for these samples (Figure \ref{fig:boxplots}) and the investigated methods were compared using their bootstrapped $1-\alpha$ confidence intervals.\\
For the voxel-based anayses using the bootstrap method to a sufficient accuracy was computationally intractable. Therefore, we computed AP and AUC over the voxels of small balanced subsets (1 healthy, 1 pathological, 44 subsets). The resulting metric values were tested for significant differences between the methods with the Wilcoxon signed-rank test. The $\alpha$-significance threshold was globally corrected for multiple testing using the conservative Bonferroni correction for 24 statistical tests, leading to an $\alpha\textsuperscript{*}=0.05/24=0.002$ threshold of significance \citep{wenz2015image}.

The investigated methods were qualitatively evaluated on clinical use cases comprising brain atrophy, tumors (large and small metastases) and movement artefacts, which are typical for neuroradiological reporting together with the most suitable sequences for diagnosing the respective pathology in Figures 2-5. The windowing levels of the figures were automatically set with the default method implemented in ITK-SNAP \citep{py06nimg}.

\section{Results}
\subsection{Quantitative evaluation}
\begin{figure}[h] 
    \centering
    \includegraphics[scale=0.70]{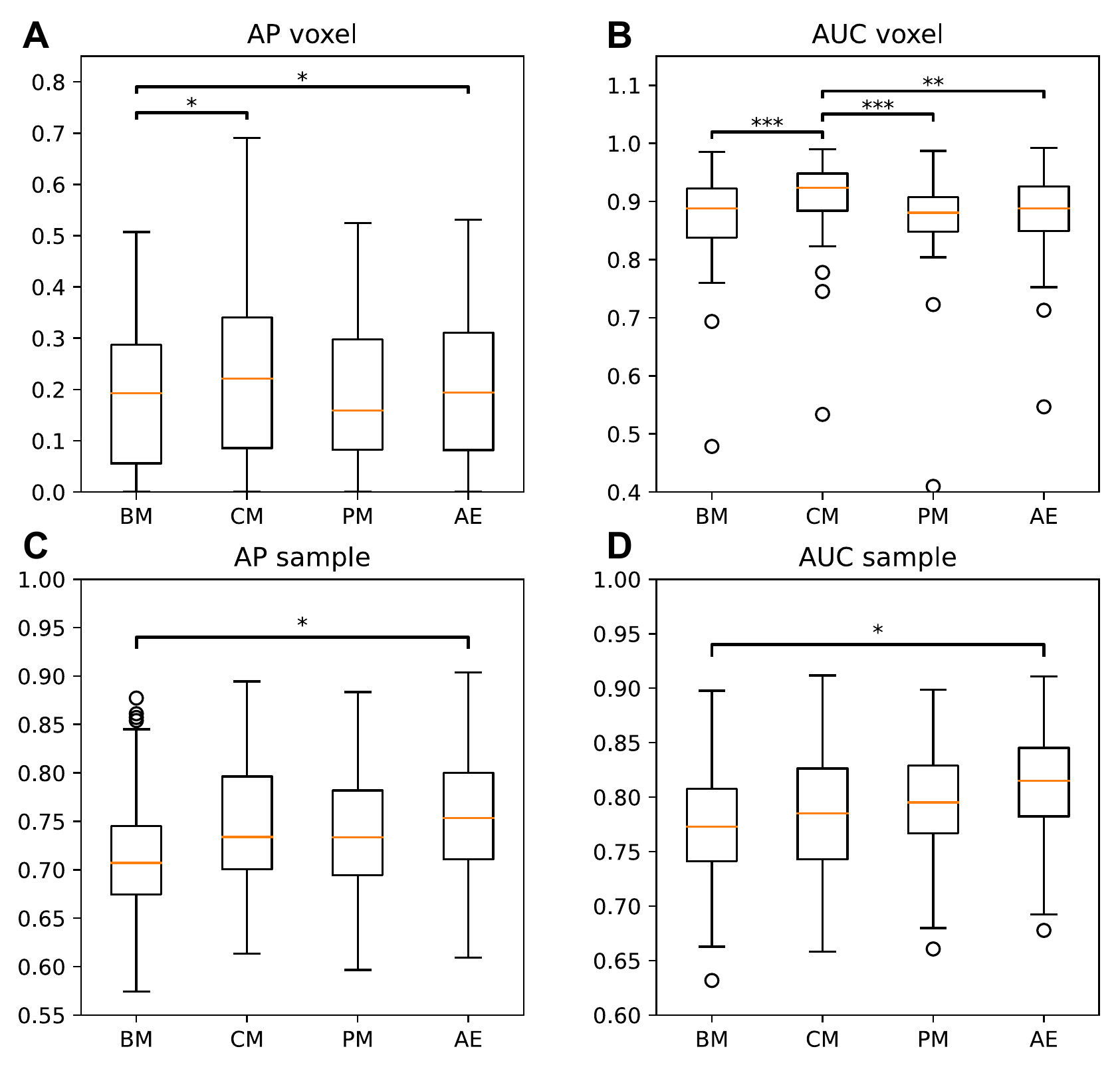}
    \caption{Boxplots of evaluation metrics. A \& B voxel-based comparative analyses using pairs of healthy and pathological cases. C \& D sample-based comparisons using 100000 bootstrapped samples. AP: average arecision. AUC: area under the ROC curve. BM: baseline model. CM: covariance model. PM: projection method. AE: autoencoder. Wilcoxon signed-rank tests or bootstrap confidence intervals of differences between the methods were marked as significant after Bonferroni correction ($\alpha = 0.05/24 \approx 0.0021$) with significance levels *: $p<0.0021$, **: $p<0.0001$, ***: $p<0.00001$. }
    \label{fig:boxplots}
\end{figure}

The results of the quantitative analyses are presented in Figure \ref{fig:boxplots} and Table \ref{table:2}. In the voxel-based task, we found that both CM (p=0.0004) and AE (p=0.0007) were significantly better than the BM, but there was no significant difference between these methods regarding AP (Fig. \ref{fig:boxplots}A). CM had a significantly better AUC than all other methods (Fig. \ref{fig:boxplots}B) while AE showed no strictly significant improvement (p=0.01) over the BM. Notably, the PM was unable to significantly improve on the AP and AUC of the BM. \\
In the sample-based task, the CM and PM showed no relevant differences while the AE showed a significant improvement (p=0.0009) over the AP (Fig. \ref{fig:boxplots}C) and AUC (Fig. \ref{fig:boxplots}D) of the baseline. We elaborate on this discrepancy between voxel-based and sample-based tasks in the \ref{Discussion} Discussion.

\begin{table}[h] 
\centering
\begin{tabular}{ l l l l l}
 Method & AP voxel & AUC voxel & AP sample & AUC sample \\
 \hline  \\ 
 BM: Baseline & 0.192 & 0.867  & 0.704  & 0.765  \\ 
 CM: Covariance & \textbf{0.235} & \textbf{0.904}& 0.732  & 0.776 \\  
 PM: Projection/PCA & 0.197 & 0.868 & 0.729  & 0.788 \\
 AE: DL Autoencoder & 0.208 & 0.876 & \textbf{0.751}  & \textbf{0.805} \\
 \\
\end{tabular}
\label{table:2}
\caption{Evaluation metrics for 44 healthy and 44 manually segmented pathological studies in the voxel-based and sample-based tasks. CM had the best voxel-based metrics while AE achieved the highest scores on the sample-based tasks. For the voxel-based task, we reported the bootstrapped median values of the metrics. AP: average precision, AUC: area under the ROC curve.}
\end{table}

\subsection{Qualitative evaluation with illustrative examples}
\begin{figure}[H]
    \centering
    \includegraphics[scale=0.25]{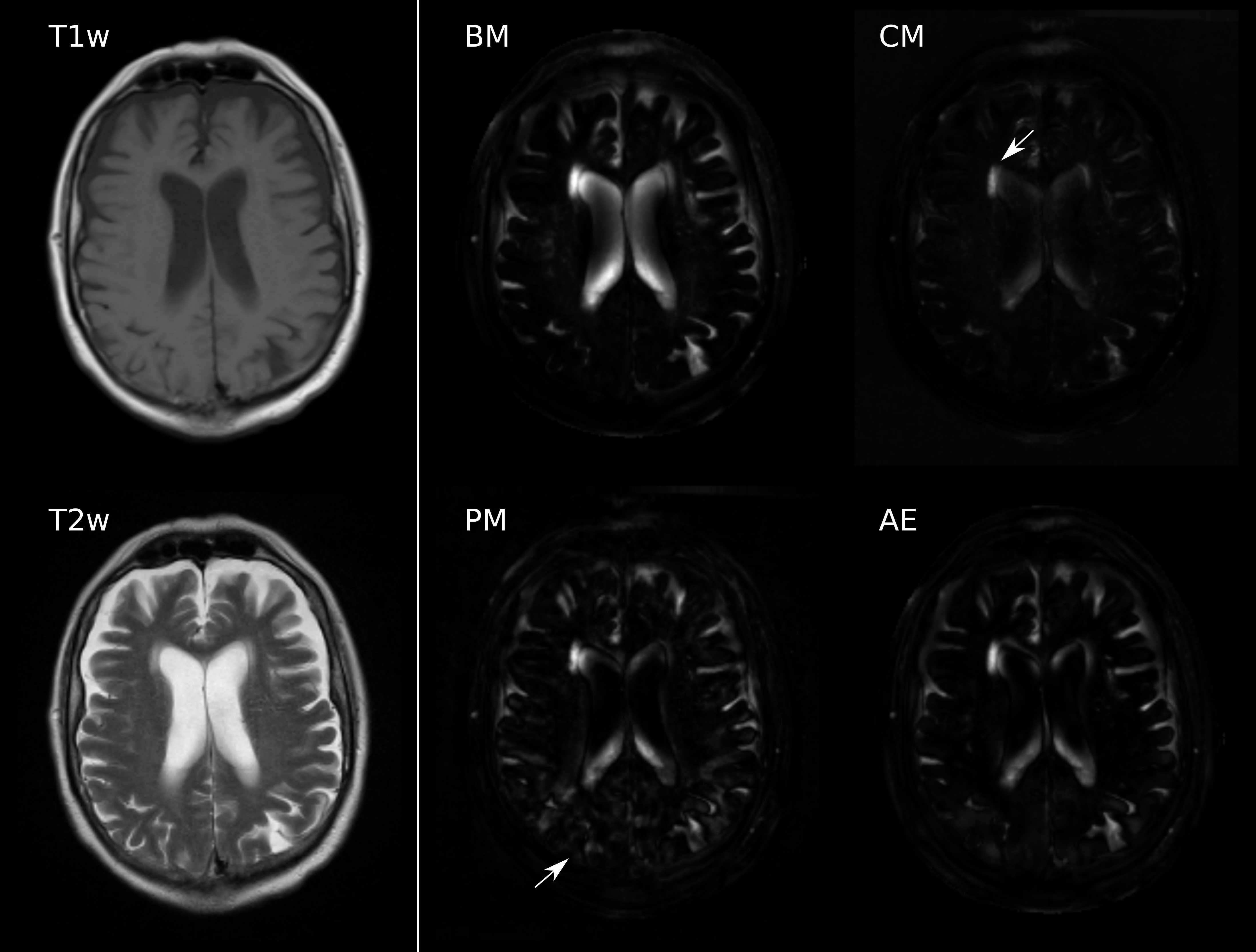}
    \caption{Brain atrophy in aging. The basic model (BM) was particularly susceptible to enlarged ventricles, while the other methods "learned" to accept the high variance of ventricular anatomy. The covariance method (CM) achieved the best differentiation between the physiologically enlarged right frontal horn and the microangiopathic adjacent lesions (\colorbox{Black}{\textcolor{white}{$\leftarrow$}}). It also correctly ignored most of the physiologically enlarged outer subarachnoid spaces. The projection method (PM) introduced new diffuse false positive artefacts in the occipital regions (\colorbox{Black}{\textcolor{white}{$\leftarrow$}}). The autoencoder (AE) performed between the baseline and covariance methods regarding the right ventricule and adjacent white matter lesion detection and correctly identified the enlarged left occipital sulcus.}
    
    \label{fig:img_csf}
\end{figure}

\begin{figure}[H]
    \centering
    \includegraphics[scale=0.25]{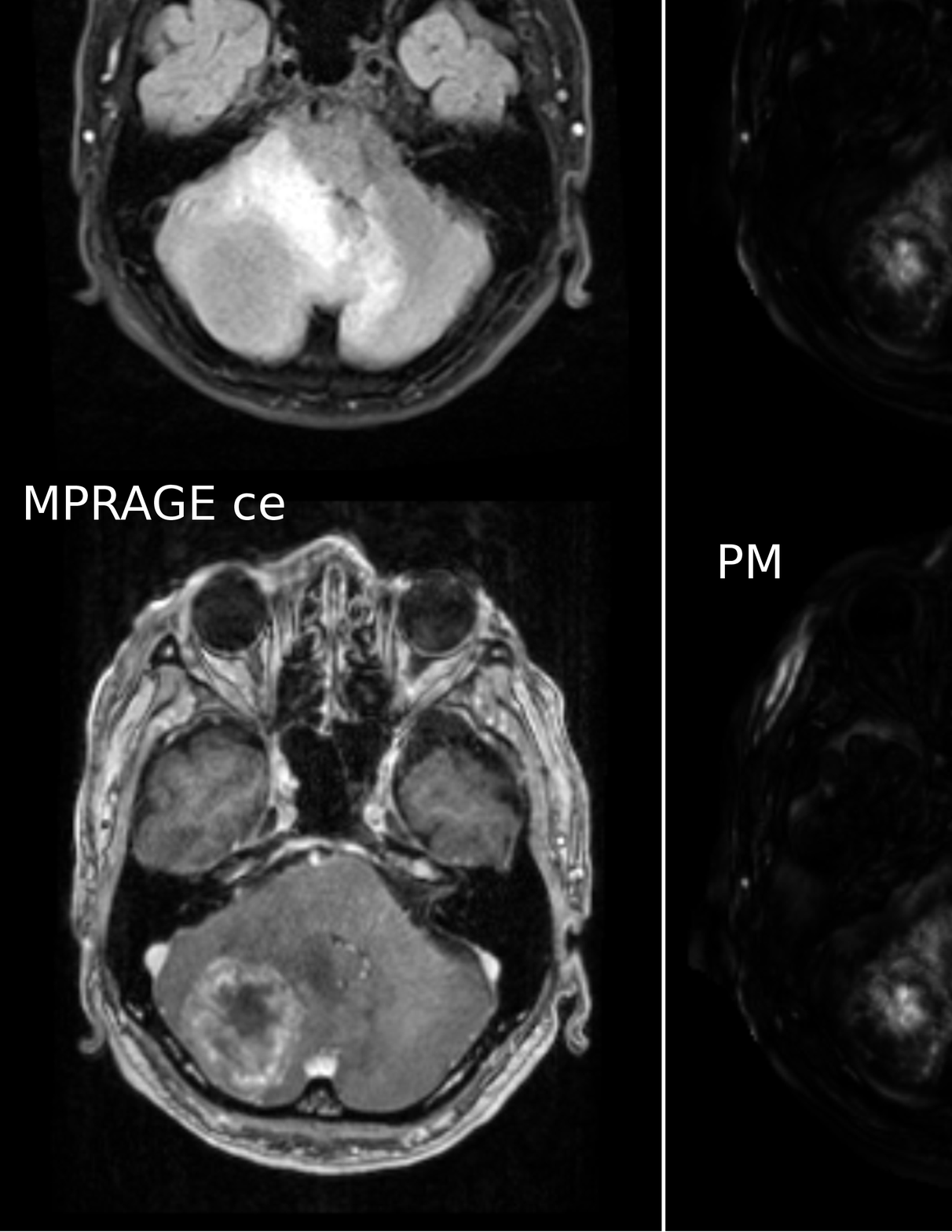}
    \caption{Contrast enhancing solitary brain tumor with mass effect and perifocal edema. The covariance model (CM) captured more details of the heterogeneous, rim enhancing tumor portion as well as the perifocal edema (\colorbox{Black}{\textcolor{white}{$\leftarrow$}}). In this case, also the autoencoder (AE) improved on the basic model (BM). It was especially suitable for detecting the necrotic core of the tumor. All models correctly identified the extracranial right periorbital subcutaneous swelling.}
    \label{fig:img_tumor2}
\end{figure}

\begin{figure}[H]
    \centering
    \includegraphics[scale=0.25]{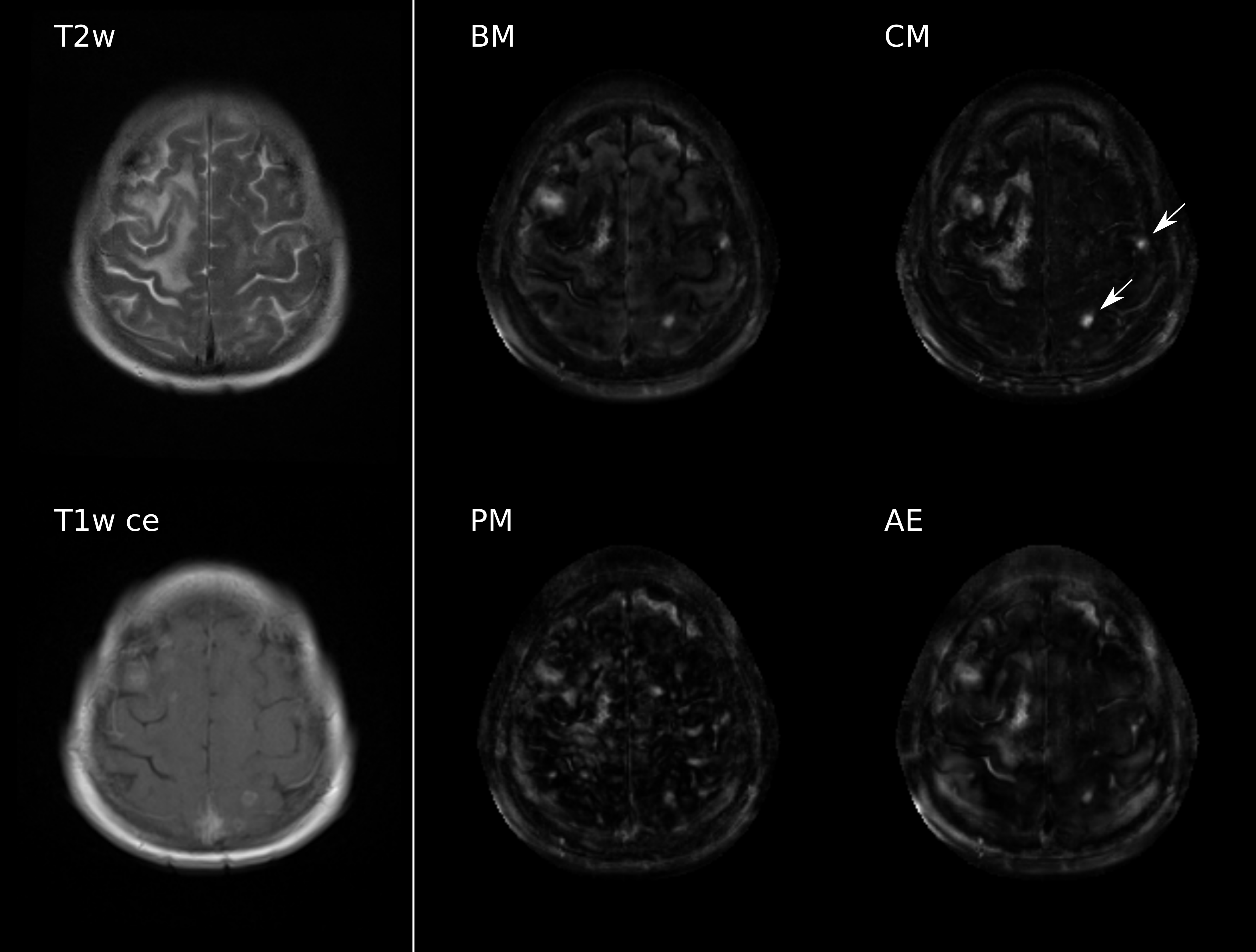}
    \caption{Multiple diffuse metastases. This case of diffuse small metastases illustrated the problems arising with spatial models like the projection method (PM) and the autoencoder (AE). They both loose spatial accuracy for detecting small lesions like the two metastases (\colorbox{Black}{\textcolor{white}{$\leftarrow$}}) in the left hemisphere within the white matter and cortical near the sulcus. The covariance method (CM) showed the best overall contrast between lesions, edema and healthy regions.}
    \label{fig:img_met}
\end{figure}

\begin{figure}[H]
    \centering
    \includegraphics[scale=0.25]{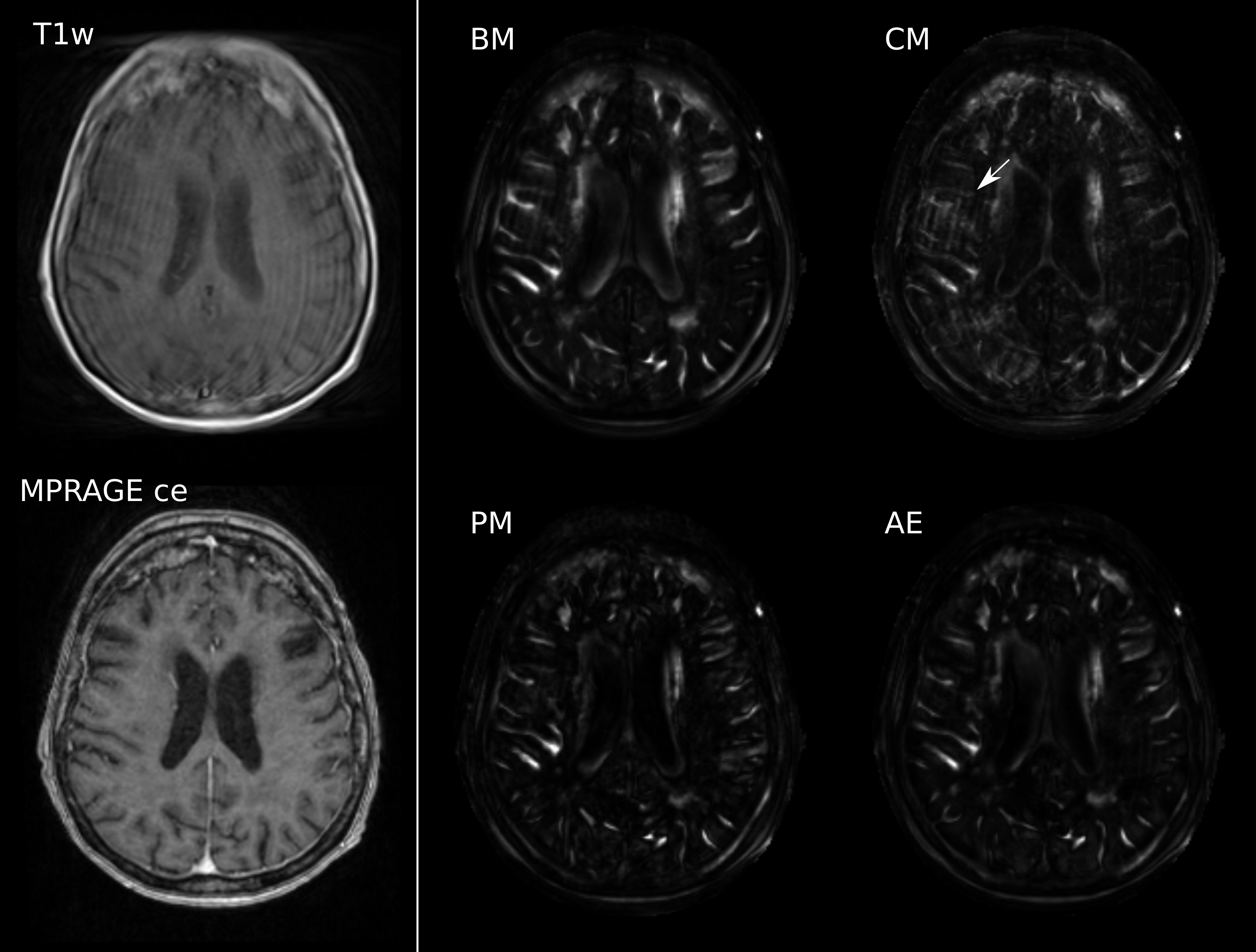}
    \label{fig:img_movement}
    \caption{Movement artefacts. A strength of the spatial model including the projection method (PM) and the autoencoder (AE) was their robustness against movement artefacts. Here, the covariance model (CM) enhanced the artefacts in comparison to the basic model (BM) (\colorbox{Black}{\textcolor{white}{$\leftarrow$}}) while PM and AE substantially reduced artefacts compared to both CM and BM.}
\end{figure}

Qualitative evaluation of the methods were performed using the resulting residual maps and presented in Figures 3-5 depicting the most suitable sequences for diagnosing the respective type of pathology by neuroradiologists including brain atrophy (Fig \ref{fig:img_csf}), large solitary primary brain tumors (Fig \ref{fig:img_tumor2}), multiple small metastases (Fig \ref{fig:img_met}) or movement artefacts (Fig \ref{fig:img_movement}). \\
The large (partly age-dependent) variability of the anatomy of the ventricular system and the external cerebrospinal fluid spaces (sulci and cisterns) posed a great challenge for most of the models (Figure \ref{fig:img_csf}). Overall, CM performed the best on this task while BM was particularly vulnerable for enlarged ventricles, PM falsely introduced sulcal artefacts and AE performed between BM and CM when differentiating between periventricular white matter lesions and ventricular enlargement. \\
When detecting large solitary tumors with mass effects (Figure \ref{fig:img_tumor2}), it was noticeable that the contrast enhancement in regions of usually intact blood-brain barrier was more strongly visible in CM compared to BM. This effect was due to the high covariance between non-contrast T1 and contrast enhanced MPRAGE in the healthy case that enabled the model to better detect deviations from this covariance in the case of pathological enhancement. \\
Fundamental differences became apparent in the ability of the investigated models to concurrently detect small metastases while being able to discard movement artefacts. The case of diffuse small metastases (Fig. \ref{fig:img_met}) illustrated the problem of spatial models (PM and AE) losing fine spatial accuracy thereby missing small lesions. On the other hand, a particular strength of PM and AE seemed to be that they were less susceptible to movement artefacts (Fig. \ref{fig:img_movement}), which in contrast to small lesions show long-range spatial correlations.

\section{Discussion}
\label{Discussion}

We performed a comprehensive comparison of algorithms from traditional linear voxel-wise statistics without (baseline) and with covariance structure (CM) over spatial projections (PM/PCA) to deep learning (convolutional autoencoder) for the unsupervised detection of general pathologies in multi-parametric head MRI studies. The investigated methods were trained on a novel, expert-curated database of healthy MRI studies, hence they were not specific to certain kinds of pathologies and their distributions \citep{Zimmerer2018ceVAE, Zimmerer2019, Maros2020machine}. We demonstrated through both voxel- and sample-based quantitative and also qualitative comparisons that while the AE performed well, also the much simpler CM produced competitive results. Particularly for spatial sensitivity and the detection of contrast enhancement, CM was superior to the 3D convolutional AE, as demonstrated on the case of small metastases. In contrast, however, only the AE showed superior results to the baseline model in the sample-based task. 

This discrepancy between voxel- and sample-based performance could be explained by the spatial correlations that the convolutional DL architecture is designed to learn \citep{Baur2020, Baur2020b, Zimmerer2018ceVAE, Zimmerer2019}. Pathological voxels lead to deviations in the predictions also for neighboring healthy voxels, thereby losing spatial accuracy but gaining sensitivity to the general presence of pathologies. Spatial models might also better in correcting spatial artefacts, as shown in the case of movement artefacts. DL autoencoders can be understood as a nonlinear generalization of PCA \citep{bourlard1988auto}. Therefore, we also investigated the performance of a linear projection method that is equivalent to PCA. Accordingly, our results illustrated a loss of spatial accuracy for both of these methods. 

When using DL methods, one has to be attentive to certain known limitations coming with them, such as their black box nature and their poorly understood susceptibility to variations in the training and testing data \citep{Biggio_2018}. The output of these algorithms needs to be interpreted by radiologists, therefore, the algorithms and their outputs must to some degree be comprehensible for human experts \citep{Geis2019}. 
Nonetheless, DL approaches offer a very flexible framework with many degrees of freedom in the specification of the models \citep{SENGUPTA2020105596}. We would not be surprised if some architectures showed a significantly improved precision over the models presented here. However, also our voxel-wise models may easily be extended to (e.g. hierarchical linear or nonlinear regression) models, which take into account covariates like MRI parameters in principled ways \citep{gelman:2006}. Physically grounded models based on MRI acquisition physics and quantitative measurements may be another worthwhile direction to pursue before resorting to black-box DL methods \citep{west2012novel}. 

Other than DL, also clustering-based methods have been proposed for the unsupervised detection of pathologies in brain MRI \citep{Juan-Albarracin2015}. Clustering-based models estimate tissue types by taking into account all voxels and then mark voxels as pathological, if they do not fit well into the known tissue classes. This is also a common approach for automatic segmentation of the brain into white matter, gray matter and corticospinal fluid \citep{arnaud2018fully, ji2012fuzzy}. The voxel-wise covariance model presented here can be interpreted as a clustering method with only a single cluster, parametrized by voxel location. This is also known as one-class classification. This can be considered one of the most simple unsupervised methods for anomaly, novelty or out-of-distribution detection. We showed that it can perform better than DL methods.

Our imaging cohort was comprised of not only more patients than the frequently used BraTS 2018 dataset (N=439 vs. N=243) but also included more sequences (9 vs. 4) \citep{bakas2017advancing,bakas2019identifying,menze2014multimodal}. The MICCAI MOOD 2020 challenge for unsupervised brain pathology detection provided a larger cohort of 800 cases, although limited to only contrast enhanced MPRAGE sequences \citep{mood_2020_3784230}. On the MOOD 2020 challenge, the PM/PCA method achieved the 5th best pixel-level score (AP 0.204) and 6th best sample-level score (AP 0.800) out of 8 models. Notably, it was in the range of highly complex, mostly AE-based DL methods (\url{https://www.youtube.com/watch?v=yOemj1TQfZU}), which considering its faster training and direct interpretability makes PM/PCA an appealing choice for computer-assisted diagnostic tools \citep{Maros2020MLradlexCAD}. 
Regarding the evaluation, there is an important difference between the applied performance metrics, as AP is better suited to evaluate the ability of a method to correctly detect pathologies within an image slice while AUC describes the ability to correctly classify both healthy and pathological voxels \citep{davis2006}. This distinction is important, because usually only a small percentage of the voxels in pathological MRI scans are classified as pathological \citep{mood_2020_3784230,bakas2017advancing,bakas2019identifying,menze2014multimodal}. Hence leading to a large AUC value even for an algorithm that marks every voxel as healthy. Accordingly, the large overlap of the boxplot whiskers in Figure \ref{fig:boxplots} can be explained by the variance of the samples in the testing set, which is grounded in the varying difficulty of detecting the differing kinds of pathologies. Similarly, the noticeably overall lower AP and higher AUC values in the voxel-based task compared to the sample-based task are due to the imbalanced distribution of pathological vs. healthy voxels with only about 3\% of the voxels being pathological in a typical pathological sample in our data set \citep{mood_2020_3784230,bakas2017advancing}. 

Our study had certain limitations, as we did not strive to beat the state-of-the-art DL model but to evaluate a reasonable and well-established architecture, which is similar to what was presented in other studies \citep{Baur2019,Baur2020}.
We acknowledge that not using a separate validation set to determine the end of the AE training period might be considered suboptimal. Nonetheless, we did not observe overfitting (see suppl. Figure \ref{fig:supplconvergence}) and the potential amount of information leakage was very low. It is of note, however, that we evaluated the model at different epochs and using the last epoch led to only marginally worse performance metrics. Additionally, we deliberately investigated only naturally occurring anomalies for clinical applications and did not attempt to comprehensively evaluate the robustness of these methods to synthetic perturbances like in the MICCAI MOOD 2020 challenge \citep{mood_2020_3784230}. 

In conclusion, we demonstrated that a frequently proposed deep learning approach (3D convolutional autoencoder) for unsupervised detection of pathologies on brain MRI was not able to improve significantly on a simple voxel-wise model in detecting lesions. Therefore, we advocate that simple methods should be used as baseline models for comparison when publishing DL-based algorithms in unsupervised anomaly detection. Our results should also serve as a note of caution to clinical practitioners in the process of adopting DL-methods into their clinical workflow.

\newpage

\section{Data statement}
The curated imaging datasets analyzed during the current study are available from the authors on reasonable request. Source code is available on GitHub (\url{https://github.com/vsaase/simpleBAD}) and trained models are shared on Mendeley Data.  

\section{Acknowledgements}
V.S. and M.E.M. were supported by funding from the German Federal Ministry for Economic Affairs and Energy, Zentrales Innovationsprogramm Mittelstand (ZF 4514602TS8). M.E.M. and T.G. acknowledge funding from the German Ministry for Education and Research within the framework of the Medical Informatics Initiative (MIRACUM Consortium: Medical Informatics for Research and Care in University Medicine; 01ZZ1801E).

\section{Author Contributions}
\textbf{Victor Saase}: Conceptualization, Methodology, Software, Validation, Formal analysis, Investigation, Data Curation, Writing - Original Draft, Visualization,
\textbf{Holger Wenz}: Conceptualization, Validation, Writing - Review \& Editing, 
\textbf{Thomas Ganslandt}: Resources, Writing - Review \& Editing,
\textbf{Christoph Groden}: Resources, Writing - Review \& Editing,
\textbf{Máté E. Maros}: Conceptualization, Methodology, Validation, Formal analysis, Visualization, Writing - Original Draft, Writing - Review \& Editing, Funding acquisition, Supervision.
All authors read and approved the final version of the manuscript.

\section{Conflict of interest}
V.S. and M.E.M. received funding from the German Federal Ministry for Economic Affairs and Energy, Zentrales Innovationsprogramm Mittelstand (ZF 4514602TS8). M.E.M and T.G. received funding from the German Ministry for Education and Research within the framework of the Medical Informatics Initiative (MI-I 01ZZ1801E). The founding sponsors had no role in the design of the study; in the collection, analyses, or interpretation of data; in the writing of the manuscript; and in the decision to publish the results. The other authors declare no conflicts of interest.

\newpage

\bibliography{references}  

\appendix
\section{Supplementary material}

\begin{figure}[H]
    \centering
    \includegraphics[scale=0.5]{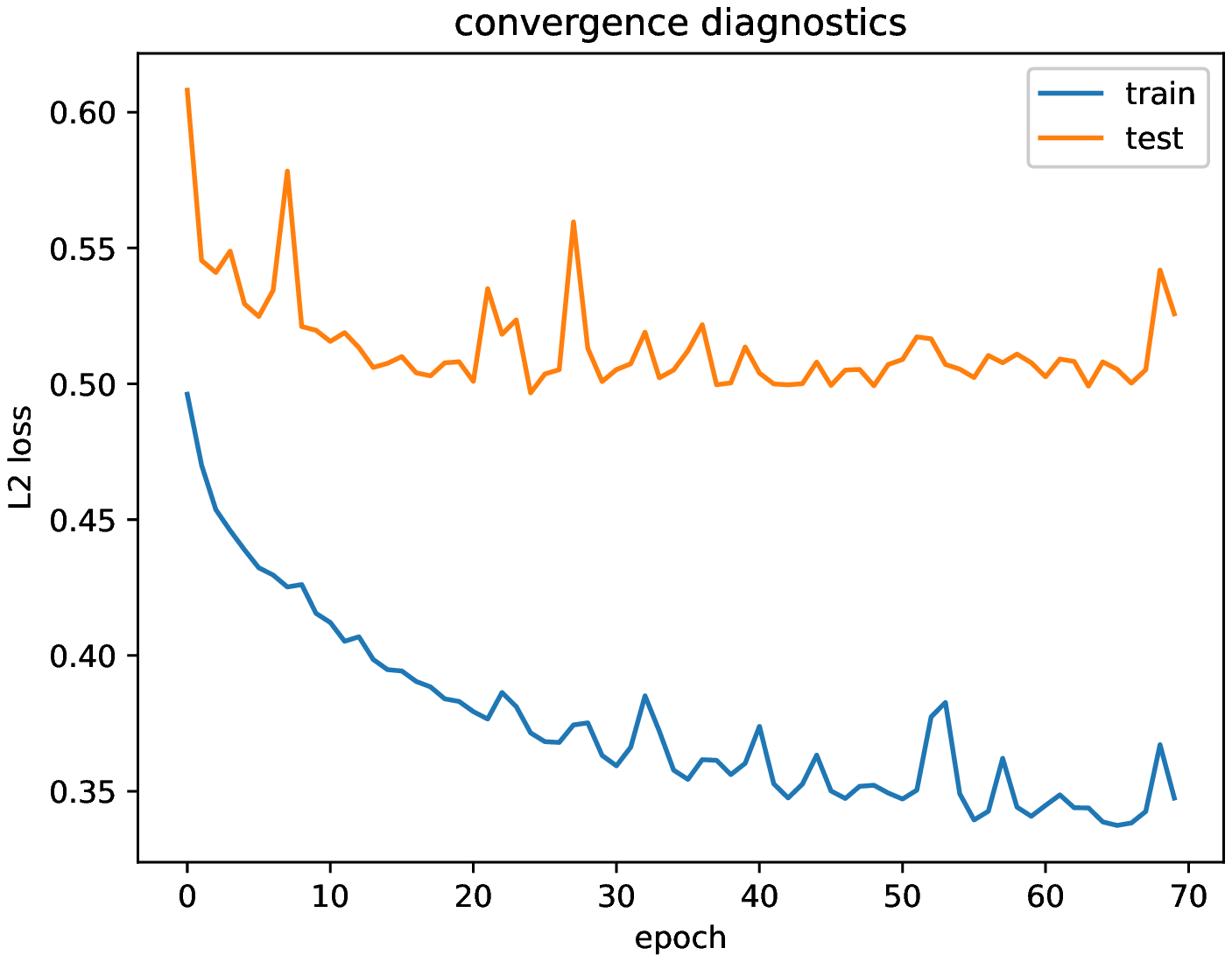}
    \caption{Convergence diagnostics of the autoencoder. We observed no overfitting and no improvement of the test error after epoch 25. Training was stopped after a maximum of 70 epochs. The selected final model was the one with the lowest testing/validation loss.}
    \label{fig:supplconvergence}
\end{figure}
\end{document}